\documentclass{article}

\usepackage[preprint]{neurips_2026}

\usepackage[utf8]{inputenc}
\usepackage[T1]{fontenc}
\usepackage{hyperref}
\usepackage{url}
\usepackage{booktabs}
\usepackage{xcolor}
\usepackage{array}
\usepackage{graphicx}
\usepackage{amsmath,amssymb,amsfonts}
\usepackage{microtype}
\usepackage{caption}

\title{DungeonBench: A Benchmark for Rules-Rich Tactical Reasoning in Dungeons \& Dragons Combat}

\author{%
Ismayil Ismayilov \\
fal \\
\And
Atakan Kara \\
fal \\
\And
Kaan Oktay \\
fal \\
}

\begin{document}

\maketitle

\begin{abstract}
Games and simulators make valuable benchmarks by turning decisions into measurable outcomes, but many current suites under-test rules-rich tactical reasoning: the ability to choose well when geometry, timing, resources, objectives, and rule interactions all matter at once. We introduce \emph{DungeonBench}, a benchmark for tactical reasoning in Dungeons \& Dragons combat, built to cover the vast majority of combat-relevant 2014 System Reference Document content whose effects can be resolved by the simulator while retaining mechanics that simplified combat simulators often abstract away. At each step, DungeonBench exposes a complete tactical observation, a pending decision, and an indexed list of executable options spanning movement, attacks, spells, reactions, objectives, preparation, and scarce resources. The task is to value legal choices whose consequences depend on action economy, creature traits, battlefield geometry, timing windows, and future encounters. DungeonBench has two tracks: \emph{Encounter}, which evaluates local tactical play in single fights, and \emph{Day}, which links encounters through persistent hit points, spell slots, consumables, preparation, and short-rest timing, forcing policies to trade off immediate tactical advantage against future survivability. The same engine-generated decision stream supports heuristic controllers, language-model policies, learned option rankers, and masked-action reinforcement-learning agents. We evaluate frontier language-model policies on this shared decision stream. Results show that full tactical observations do not saturate the benchmark: frontier policies often win direct encounters, but linked encounter days expose failures in resource budgeting, rest timing, and rule-aware tactical discipline.
\end{abstract}

\section{Introduction}
\begin{figure}[htbp]
    \centering
    \includegraphics[width=1.0\textwidth]{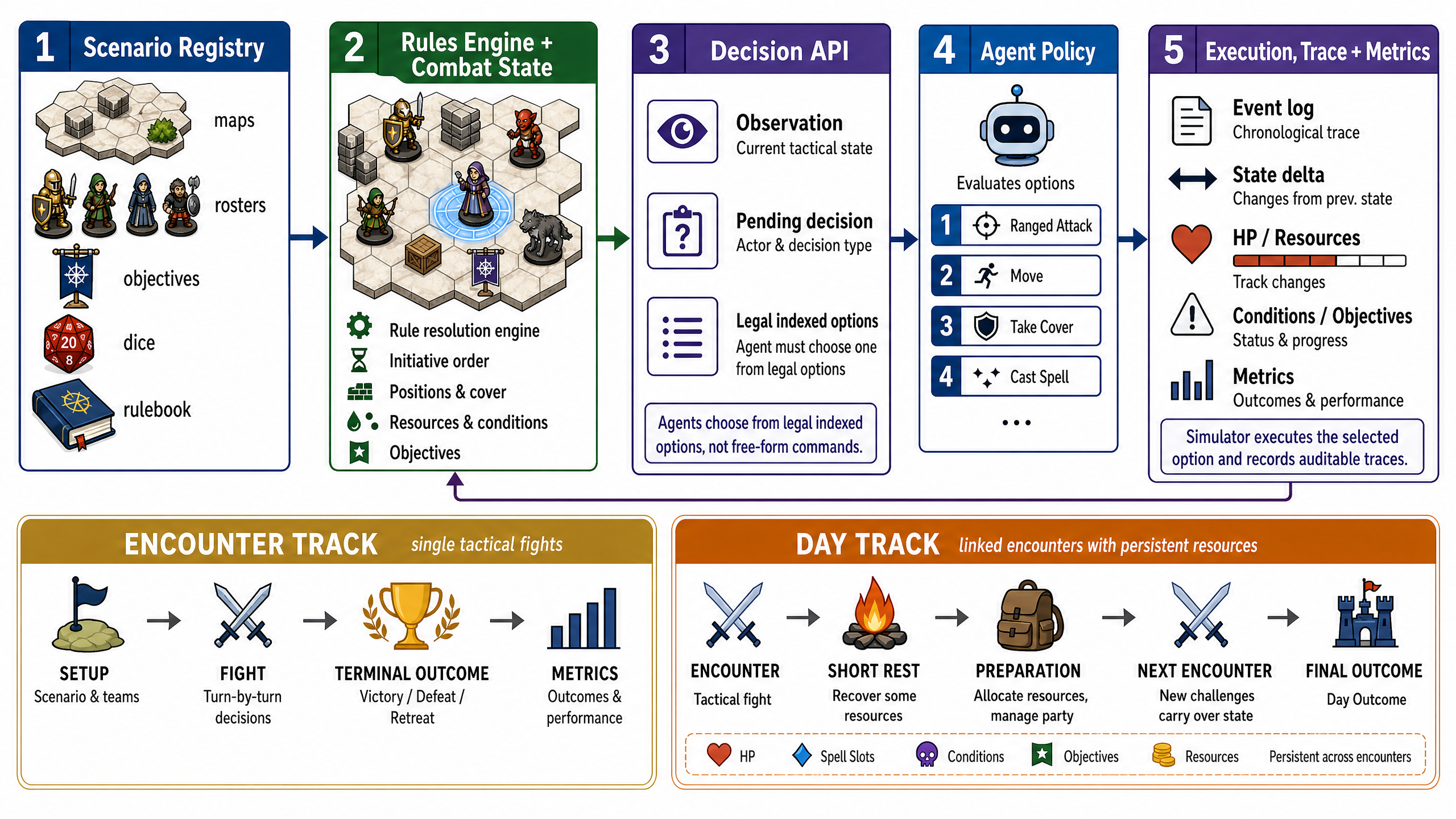}
    \caption{Overview of DungeonBench. The engine maps tactical state to legal executable options; policies choose over the same decision stream, and the simulator scores Encounter and Day outcomes.}
    \label{fig:dungeonbench}
\end{figure}

Games make robust testbeds for artificial intelligence: explicit rules, measurable outcomes, and repeatable interaction loops give an unambiguous account of what an agent did and what its decision produced. As a result of this, they have supported progress in continuous control, board games, multi-agent sports, procedural navigation, and open-ended worlds \citep{brockman2016openai,cobbe2020procgen,lanctot2019openspiel,zhang2025volleybots}. Yet one kind of reasoning remains underrepresented: tactical judgment in domains where legal actions are heterogeneous and their value is determined by rule interactions, timing windows, objectives, and scarce resources.

In a tactical domain, one decision may be about where to stand, another about which enemy to target, another about whether to spend a scarce spell slot, trigger a reaction, protect concentration, exploit terrain, or hold a resource for a later fight. These choices draw on different action types and acquire value through the rules around them. A locally strong action can be strategically wrong; a low-damage move can be decisive because it preserves concentration, blocks a doorway, saves a reaction, or keeps a resource for the next encounter. The challenge is not only long horizon or large branching factor. It is state-contingent option valuation under executable rules.


The source domain is the combat-relevant portion of 2014 Dungeons \& Dragons. In Dungeons \& Dragons combat, parties and opposing creatures act in initiative order on a battlefield; turns combine movement, attacks, spells, object interactions, reactions, and resource spending. We use this domain because it offers a mature library of tactical primitives rather than a rule system invented solely for the benchmark. Its mechanics have been accumulated, revised, and stress-tested through years of play, giving DungeonBench a broad rule surface where positioning, resources, timing, and rule interactions all matter without being tuned to any particular model class. We restrict attention to combat-relevant rules that can be represented and resolved by the engine.

DungeonBench turns this problem into a benchmark for tactical reasoning. It exposes each decision as a complete tactical state together with an indexed list of legal options generated by the rules engine. Agents therefore do not need to produce command syntax or discover the legal action set; they must rank legal choices of different kinds whose consequences are resolved by the simulator as illustrated in \ref{fig:dungeonbench}.

The benchmark has two tracks. The \emph{Encounter} track tests local tactical play in single fights: positioning, target priority, reactions, spell interactions, party coordination, boss mechanics, and objectives. The \emph{Day} track links encounters through persistent party state, scarce resources, preparation choices, and short rests. This is where ``win the fight'' and ``win the day'' diverge. A high-level spell, an ability, a hit die, or a consumable can be correct to spend now if it ends a fight, or wrong if it leaves the party unable to answer a later room. DungeonBench makes that pressure visible and measurable by carrying party state, resources, active effects, and rest choices across encounters rather than treating each battle as an isolated episode.

We make three contributions. \textbf{(i)} We introduce \textbf{DungeonBench}, a combat-rules engine and benchmark for reasoning over the 2014 D\&D tactical kernel, in which agents rank state-dependent options enumerated by the engine. \textbf{(ii)} We introduce \textbf{Encounter} and \textbf{Day} tracks that test local tactics, reactions, spell and item semantics, boss mechanics, objectives, preparation, and medium-horizon resource budgeting. \textbf{(iii)} We provide direct heuristic, text-prompt, tensorized, and Gymnasium-compatible masked-action interfaces over the same engine-generated decision stream, and use this shared surface to evaluate frontier language-model policies.

\section{Related Work}

\textbf{RL environments and benchmark suites.}
OpenAI Gym and successor interfaces established a common pattern for environment interaction \citep{brockman2016openai}. PettingZoo extends related abstractions to multi-agent environments \citep{terry2021pettingzoo}, while OpenSpiel collects game-theoretic environments and algorithms \citep{lanctot2019openspiel}. Procgen emphasizes generalization across generated levels \citep{cobbe2020procgen}, and Jumanji provides scalable environments for combinatorial decision problems \citep{bonnet2024jumanji}. These suites provide reproducible interaction protocols and comparable agent interfaces across broad classes of tasks. DungeonBench follows this benchmark tradition, but centers evaluation on rule-mediated tactical option valuation.

Capability-focused game benchmarks show how a game environment can be useful because it isolates a particular form of reasoning. Hanabi foregrounds belief, intent, and ad-hoc cooperation \citep{bard2020hanabi}. DungeonBench is analogous in spirit, but isolates tactical choice under structured combat rules rather than hidden-information cooperation.

\textbf{Tactical and symbolic planning benchmarks.} SMAC and SMACv2 evaluate multi-agent tactical micromanagement in StarCraft scenarios \citep{samvelyan2019smac,ellis2023smacv2}, and MicroRTS provides a lightweight real-time strategy benchmark \citep{huang2021microrts}. MiniGrid and Boxoban/Sokoban-style domains provide compact planning problems with controllable structure \citep{chevalier2018minigrid,guez2019boxoban}. PDDLGym bridges reinforcement learning and classical planning with relational observations and dynamically valid actions from PDDL domains \citep{silver2020pddlgym}. NetHack, MiniHack, Crafter, Craftax, and MineRL expose agents to large combinatorial state spaces and long horizons \citep{kuttler2020nethack,samvelyan2021minihack,hafner2021crafter,matthews2024craftax,guss2019minerl}. The Factorio Learning Environment evaluates long-horizon resource optimization through factory construction in a procedurally generated world, with task difficulty scaling through automation depth \citep{hopkins2025factorio}.

\textbf{D\&D combat simulation.} FIREBALL and CALYPSO show that D\&D is a fertile AI domain because structured game state, natural-language play, and human tabletop workflows meet in one setting. FIREBALL collects actual-play sessions with structured game state for language and command-generation tasks \citep{zhu2023fireball}, while CALYPSO studies LLM interfaces for dungeon masters \citep{zhu2023calypso}. Recent D\&D combat work includes an RL environment with LLM-controlled adversaries in restricted scenarios \citep{dayo2025dndrl} and NTRL, which uses reinforcement learning for encounter generation and dynamic difficulty adjustment rather than agent tactical control \citep{romeo2025ntrl}.

\textbf{Language models as decision policies.} Frontier language models are increasingly used as policies over structured choices: they read state, deliberate over candidates, and return a selected action \citep{yao2023reactsynergizingreasoningacting,wang2023voyageropenendedembodiedagent,wu2024smartplaybenchmarkllmsintelligent,paglieri2025balrogbenchmarkingagenticllm}. 

DungeonBench differs from prior environments by focusing on tactical control over a broad executable combat surface. It deliberately retains mechanics that simplified combat benchmarks often abstract away: level-1-to-20 level class and subclass features, multi-classing, conditions, boss mechanics, and encounter-to-encounter resource persistence. These mechanics are not edge clutter; they are the interactions that make tactical option valuation difficult.

\section{Benchmark Design}

\subsection{D\&D combat background}
Dungeons \& Dragons is a tabletop role-playing game whose combat rules specify how player characters, monsters, spells, equipment, terrain, and battlefield effects interact. DungeonBench is grounded in the 2014 System Reference Document (SRD), the publicly available rules corpus for that edition. The 2014 SRD includes twelve base classes, with subclasses adding further rule packages. These choices are mechanically consequential: a fighter, cleric, rogue, and wizard differ not only in theme or statistics, but in the actions they can take, resources they spend, and tactical problems they create. Monsters are defined by stat blocks with hit points, defenses, actions, traits, reactions, and sometimes special mechanics such as legendary actions.

D\&D combat is organized into rounds, with creatures acting in initiative order. A turn is not a single atomic action: a creature may split movement around an action, use a bonus action when available, interact with objects or terrain, and trigger reactions from other creatures. Tactical state also persists across turns and encounters through hit points, spell slots, class-feature uses, item charges, conditions, concentration effects, and rests. A short rest represents about one hour and restores some, but not all, resources. Thus combat choices are not only immediate damage choices; they depend on timing, action economy, interrupt windows, positioning, and resource allocation across future decision points.

\subsection{Executable combat scope}
DungeonBench targets the combat-relevant portion of the 2014 D\&D SRD. The design rule is simple: if a spell, feature, item, monster trait, or combat rule can matter tactically and can be represented by the engine without inventing an open-ended ruling, it belongs in scope. Effects whose primary consequence depends on dungeon-master adjudication, campaign fiction, or creature interpretation are excluded or reduced to mechanically specified cases. For example, illusion spells such as \emph{Minor Illusion} are outside the main benchmark contract when their value depends on what a creature believes. The open-ended reality-rewriting use of \emph{Wish} is likewise outside scope, while mechanically specified uses such as spell duplication or rollback-style effects are modelable.

This boundary is meant to preserve breadth, not to make the simulator small. DungeonBench includes rules that are often dropped from simplified combat environments: reactions, concentration, persistent zones, summons, legendary actions, object interactions, mounted combat, underwater combat, flight, falling, vertical range, and hard-to-place effects such as \emph{Wall of Force}. The benchmarked rule surface contains 561 combatants, including 93 spellcasters and 19 combatants with legendary actions; 229 spells; 275 items; and class, subclass, ancestry, monster, condition, and persistent-effect mechanics. These numbers characterize the executable rule surface available to the task registry rather than a claim that every tabletop use of every rule is modeled.

\subsection{Data-driven engine and battlefield}
DungeonBench is both a fixed benchmark suite and a data-driven tactical-combat engine. Combatants, maps, spells, items, objectives, persistent effects, and encounter days are authored as data and resolved by a shared rules engine. This matters because the SRD supplies the underlying rule vocabulary---classes, subclasses, spells, items, monsters, conditions, and combat procedures---but not benchmark-ready parties, tactical maps, objectives, encounter-day sequences, or complete player-character rosters. DungeonBench therefore composes player archetypes and benchmark scenarios from SRD mechanics, adding authored maps, objectives, and bosses where needed to create controlled tactical tasks. Examples include a draconic sorcerer, an epic paladin/sorcerer multiclass, objective-oriented rogues and casters, and bosses such as the Tempest Lictor and the Black Bell of Ordrune. New scenarios can be built by composing existing abilities, rosters, maps, objectives, and encounter-day structure rather than changing the benchmark interface.

The battlefield uses a 2.5D grid: positions have horizontal coordinates and height. This keeps movement and targeting executable while preserving tactical effects that depend on verticality. A flying creature can gain line of sight over cover, fall, fight above a ground target, or cast an area spell downward. Terrain can block sight or movement, provide cover, create hazards, support mounted or underwater fights, and constrain vertical movement. The point is not graphical fidelity; it is to keep height, range, flight, falling, and upward or downward area templates inside the executable rules rather than flattening them away.

\subsection{Observation and decision interface}
The main observation mode is a complete tactical observation as shown in Table \ref{fig:dungeonbench}: the agent receives a serialized state sufficient for tactical decision-making, excluding non-semantic bookkeeping. The observation includes combatant statistics, hit points, spell slots, feature uses, positions, map state, conditions, active effects, zones, summons, objectives, encounter-day state, initiative context, and the current decision. These fields determine whether a creature can hold a doorway, whether a spell can affect a target, whether a reaction is available, and whether spending a resource now changes the next fight.

Full observability is deliberate. The main benchmark asks whether a policy can use the relevant tactical facts once they are available, not whether it can infer hidden monster traits, remember a scouted room, or guess that a harder fight is coming. Hidden-information and scouting variants are natural extensions of the same interface, but the reported benchmark fixes information access so that failures reflect tactical valuation rather than perception or memory.

As mentioned earlier, D\&D combat also does not fit a single atomic action interface. A turn can interleave movement, action, bonus action, object interaction, and more movement; a creature's movement can trigger another creature's opportunity attack; a spell can trigger \emph{Counterspell}, which can itself be countered; a failed save can trigger an optional resource spend; a boss can act in another creature's turn through a legendary action. DungeonBench therefore exposes combat as a sequence of rules-generated decisions rather than asking an agent to emit a full turn script.

\subsection{Hierarchical action decomposition}
The broad rule surface creates an action-space problem. A flat action space over movement, spell choice, slot level, targets, area placement, area orientation, summon placement, reaction use, and follow-up choices would be too large to present to agents and too brittle to enumerate by hand. This is especially acute in 3D. A spell with 150 feet of range at 5 feet per tile spans 61 coordinate values along each axis in a naive centered bounding cube, or more than 226,000 candidate origins before legality filtering. Even a flat plane contains 3,721 possible origins before considering other action parameters.


Point-targeted and area-origin effects illustrate the design. A turn-action option may already distinguish \emph{Fireball} cast at different legal slot levels, but the area origin is not flattened into that same option. Instead, DungeonBench can expose point selection coordinate-wise: choose an $x$ value, then a $y$ value conditioned on $x$, then a $z$ value conditioned on $(x,y)$. The selected triple is still a legal simulator point, but the policy sees a smaller sequence of choices. This supports point-origin spells, placement effects, and summons while preserving vertical tactics.

Directional templates use the same approach. For cones, lines, and similar effects, the policy chooses a direction by selecting $dx$, then $dy$, then $dz$ from canonical legal directions. This allows effects such as \emph{Cone of Cold} to remain cones in a 3D battlefield. A flying caster can aim downward, producing a different tactical footprint than a ground-level horizontal cone, without adding a free-form geometric language or flattening the spell into a simplified radius. The tradeoff is that one tabletop action may become several benchmark decisions, but this is the mechanism that lets DungeonBench retain rule breadth while keeping every choice executable and legal.

Movement creates a different decomposition problem because its consequences can depend on the route, not only the endpoint. Moving to the same destination can provoke different opportunity attacks, cross different hazards or zones, spend different movement, or interact differently with vertical terrain. DungeonBench therefore presents movement first as a reachable destination choice, then, only when needed, as a route choice. For a selected destination, the engine groups route variants by simulator-visible consequences and removes duplicates with identical consequences. The grouping is equivalence-based: routes are merged only when they expose the same simulator-visible consequences to the policy and engine. If several distinct routes remain, the agent sees the relevant path facts and chooses among them; if only one remains, movement resolves immediately. This preserves tactical route differences without exposing every geometrically distinct path through the 3D battlefield, making flight, swimming, climbing, and other 3D movement modes viable inside the same action interface.

\subsection{Tracks and task families}
DungeonBench defines two benchmark tracks. \textbf{Encounter} tasks test local combat planning, positioning, target selection, reactions, terrain use, spell interactions, and occasional objective play. In this setting, spending a high-level resource immediately can be correct if it wins the fight. \textbf{Day} tasks link encounters through persistent party state and limited short rests, testing whether a policy can carry tactical judgment across several rooms. The same spell slot that ends a wolf skirmish quickly may be the resource needed to survive a later boss. A druid’s Natural Recovery choice after a short rest can likewise determine whether the party regains a control spell for the next room or preserves recovery capacity for a harder encounter later. The two tracks therefore separate immediate tactical competence from resource-aware tactical planning.  

Tasks combine combatants, spells, maps, objectives, encounters, and encounter-day scenarios. The suite is organized by tactical skill rather than rule-surface size: each task family isolates a different form of choice pressure while sharing the same API. Some tasks reward clean target priority; others reward countering an enemy caster, protecting concentration, timing a reaction, exploiting terrain, or declining an attractive action because the next fight matters more. Objective tasks include movement and object-interaction goals, where class features such as a thief rogue's Fast Hands can change the value of reaching and activating a scene object. The scenarios cover reaction control through broken sightlines, gaze monsters where cover matters more than damage, underwater and vertical terrain, swarms and body-control threats, possession and affliction recovery, item-driven objectives, boss and lair tempo, and linked days that combine several of these pressures behind limited rest windows.

%
\textbf{Day regimes.}
The Day track uses a known encounter-day setting in which the encounter sequence is visible to the agent. This is deliberate. The benchmark is not asking whether an agent can guess that a harder room exists; it is asking whether the agent can act on that knowledge. The agent can see the linked fights but must decide how to allocate hit points, spell slots, consumables, short rests, and long-duration effects across them. Known future pressure becomes part of present action choice, making resource budgeting an observable tactical problem rather than hidden-information inference.

\textbf{Intermission abstraction.}
Between encounters, the Day track exposes an intermission decision phase for healing, consumables, preparation actions, persistent summons, short-rest choices, and long-duration buffs. The abstraction preserves duration-relevant tactical distinctions without modeling a full dungeon clock. Very short combat-only effects expire before the next encounter; next-encounter effects carry into one subsequent fight; one-hour effects persist until a short rest or equivalent time advance; and adventuring-day effects such as long-duration buffs persist across short rests. Some resources refresh on short rests while others do not, so limited rest windows create planning pressure over hit points, hit dice, spell slots, class features, consumables, and persistent effects. This makes preparation and rest timing part of the same decision problem as combat positioning and target choice: casting a high-level \emph{Magic Missile} into an easy first encounter can be locally effective and strategically wrong. The bounded intermission model keeps episodes reproducible while preserving the resource-pressure structure of the source game.

\textbf{Reading a trajectory.}
DungeonBench keeps this tactical layer deterministic and records both combat events and day-level state changes. The following excerpts illustrate the two tracks.

\begin{quote}
\small
\textbf{Encounter example.} In Moonhook Eyrie, a ranger opens with \emph{Volley} against a wyvern and two air elementals. An elemental uses \emph{Whirlwind}, pushing Vael 20 feet and dealing 18 bludgeoning damage; Vael answers by casting \emph{Feather Fall} on Aurel, slowing the descent. Aurel's \emph{Guiding Bolt} hits the wyvern for 31 radiant damage. Vael then casts \emph{Fireball}, catching the wyvern and both elementals; on the next round, after the manticore hits Vael three times, Aurel spends a 6th-level \emph{Healing Word} to restore 30 hit points before Vael retreats and defeats the wyvern with another \emph{Fireball}.

\textbf{Day example.} Graywatch March begins with four party members and one short rest. After North Nave, Cassian is at 19/33 hp and Nix at 30/38 hp. The party spends its only short rest: Nix spends a hit die to heal to 38/38, while Cassian spends hit dice to return to 33/33. After Lower Yard, Cassian drops to 8/33 and Nix to 32/38; with no short rest left, Ilyra uses Lay on Hands before the final Ridge Room. The day is cleared, but the final party is badly depleted: Caldrin 17/44 hp, Ilyra 0/44 hp, Cassian 0/33 hp, and Nix 0/38 hp.
\end{quote}

\textbf{Scenario registry.}
Each benchmark scenario is identified by a stable ID and binds together map, side rosters, seeds, round limit, reward, decision interface, and observation mode. The headline evaluation suite contains 25 scenarios total: 20 Encounter scenarios and 5 Day scenarios. Encounter evaluation uses three seeds per scenario, while Day evaluation uses one fixed seed per day, for 65 headline evaluation episodes. Encounter scenarios vary party and enemy composition, map geometry, objectives, environmental conditions, and boss or lair mechanics; Day scenarios vary encounter sequence, short-rest pressure, preparation opportunities, objective structure, and final boss difficulty. Detailed lists of both tracks are provided at Table \ref{tab:encounter-catalog} and \ref{tab:day-encounters}.

\subsection{Shared policy surface}
The same decomposed decision stream is the shared policy surface for DungeonBench. In the reported setting, symbolic baselines and language-model policies choose among engine-generated legal options; the simulator executes the selected option, so comparisons measure tactical option valuation rather than command parsing or action validity. The released interface also exposes the same decision stream as tensorized variable-option examples and as a Gymnasium-compatible wrapper for learned rankers and reinforcement-learning agents. Appendix~\ref{app:policy-interface} documents these text and tensor interfaces.

\textbf{Language-model policies and rule lookups.}
Language-model policies use the shared interface through a structured text rendering. The prompt contains the current decision kind, actor, tactical state summary, encounter history, legal option identifiers, and a response schema requiring a legal choice ID. Because rule details can be too large to place inline at every decision, the prompt exposes visible rule and item indices and permits lookup tools for attacks, spells, features, traits, combat riders, items, conditions, and movement routes. Lookups return simulator-grounded mechanics, including route-specific consequences such as opportunity attacks, hazards, readied reactions, and zone triggers. Invalid responses are retried against the same schema, and no response is executed unless it selects a legal option.

\section{Evaluation}
\label{sec:evaluation}

The reported evaluation focuses on frontier language-model policies under a shared decision interface. In every episode, the tested model controls the benchmark party, and all opposing sides are controlled by the same heuristic planner. This fixes the adversary and isolates the question of interest: given the same complete tactical state and the same legal option set, how well does a model value executable tactical choices?

\subsection{Evaluation Protocol}
Each language-model policy receives a serialized tactical observation, the pending decision, recent encounter history, visible rule and item indices, and a list of legal option identifiers generated by the simulator. The model does not emit free-form commands. It must select one legal option identifier, and the simulator executes that option. The policy is invoked at every exposed decision point, including movement, turn actions, reactions, optional riders, target allocation, area placement, and day-track intermission choices.

Rules and route details are exposed through lookup tools rather than copied into every prompt. These tools return simulator-grounded mechanics for spells, attacks, features, traits, items, conditions, and movement routes. Route lookups can reveal movement cost, opportunity attacks, hazards, zone triggers, readied reactions, and other route-specific consequences.

The encounter-track comparison evaluates five models: GPT-5.5, Claude Opus 4.7, Gemini 3.1 Pro, Grok 4.3, and DeepSeek V4. Because linked-day episodes expose several hundred surfaced decisions and dominate evaluation time and cost, the Day-track comparison evaluates each scenario on one fixed seed and uses Claude Sonnet 4.6 as the Anthropic model in place of Claude Opus 4.7.

The Encounter track uses 20 canonical encounter scenarios and the Day track uses 5 linked encounter-day scenarios listed at Table \ref{tab:encounter-catalog} and \ref{tab:day-encounters}. Each day consists of 3 to 5 encounters with persistent party state, intermission decisions, and limited short rests. Encounter scenarios are evaluated with 3 seeds each; Day scenarios are evaluated with 1 fixed seed each. The day sequence is known to the agent. This makes resource planning an observable tactical problem rather than a hidden-information problem.

The primary Encounter metric is win rate: the party must defeat the opposition or satisfy the scenario-specific objective before a wipe or timeout. Secondary Encounter metrics report the final party hit-point fraction, survivor fraction, rounds to termination, and decisions per episode. These distinguish clean wins from depleted wins and short failures from long failures.

The primary Day metric is full-clear rate: the party must clear every encounter in the linked day. Secondary Day metrics report final party hit-point fraction, survivor fraction, decisions per episode, short-rest use, and encounters won.       

\subsection{Results}
Tables~\ref{tab:llm-summary} and~\ref{tab:day-summary} report the frontier language-model comparison on the Encounter and Day tracks. The headline result is that strong single-encounter performance does not transfer cleanly to linked encounter days. The best performer models on Encounter track clear slightly more than 80\% of episodes, while the best Day models clear only two of five days.

On the Encounter track, Gemini 3.1 Pro and GPT-5.5 are the strongest models, with win rates of $83 \pm 5$ and $82 \pm 5$, respectively. They also preserve the most party health, ending with about 0.80 final HP and 0.93 survivor fraction on average. Grok 4.3 is lower at $72 \pm 6$, and Claude Opus 4.7 and DeepSeek V4 are tied at $68 \pm 6$. Secondary metrics show that performance is not explained by tool use alone: DeepSeek V4 uses the most tools and tokens per decision but does not achieve a higher win rate, while GPT-5.5 reaches near-top success with the fewest decisions per episode among the five models.

\begin{table}[t]
\centering
\small
\caption{Frontier language-model policies on the Encounter track. The opponent side is fixed to the heuristic planner across all evaluations. Win rate is reported with one standard error over the evaluated encounter episodes. Final HP is the party hit-point fraction at encounter end, and Survivors is the fraction of party members alive at encounter end. Decisions/ep.\ is the mean number of surfaced decisions per episode. Tools/dec.\ and Tokens/dec.\ summarize lookup-tool use and prompt-plus-completion tokens per surfaced decision.}
\begin{tabular}{lrrrrrr}
\toprule
\textbf{Model} & \textbf{Win rate} & \textbf{Final HP} & \textbf{Survivors} & \textbf{Decisions/ep.} & \textbf{Tools/dec.} & \textbf{Tokens/dec.} \\
\midrule
Gemini 3.1 Pro & $83 \pm 5$ & 0.80 & 0.93 & 93 & 1.62 & 42k \\
GPT-5.5 & $82 \pm 5$ & 0.79 & 0.93 & 79 & 1.38 & 15k \\
Grok 4.3 & $72 \pm 6$ & 0.59 & 0.78 & 102 & 1.17 & 16k \\
DeepSeek V4 & $68 \pm 6$ & 0.67 & 0.82 & 104 & 3.27 & 79k \\
Claude Opus 4.7 & $68 \pm 6$ & 0.65 & 0.82 & 109 & 0.52 & 26k \\
\bottomrule
\end{tabular}
\label{tab:llm-summary}
\end{table}

The Day track is substantially harder. GPT-5.5 and Gemini 3.1 Pro each clear two of five days, Claude Sonnet 4.6 and DeepSeek V4 each clear one, and Grok 4.3 clears none. The decision horizon is also much longer: day episodes average roughly 328--391 surfaced decisions per episode, compared with 79--109 on the Encounter track. This gap is central to the benchmark. The model must not only choose locally useful actions, but also preserve hit points, spell slots, consumables, and recovery opportunities across several fights.

Final HP and survivor fractions on the Day track should be read together with full-clear rate. Some failed days end with substantial remaining party health, which indicates objective or sequencing failures rather than gradual wear-down. Conversely, some days are nearly cleared before a final wipe, showing that partial encounter progress is not equivalent to day success. The Day track therefore exposes failures that are largely invisible in isolated encounters.

\begin{table}[t]
\centering
\small
\caption{Frontier language-model policies on the Day track. Each day chains multiple encounters with persistent state, intermissions, and limited short rests. The opponent side is fixed to the heuristic planner across all evaluations. Day clear is the full-clear rate over the five evaluated day scenarios. Final HP is the party hit-point fraction at day end, and Survivors is the fraction of party members alive at day end. Decisions/ep.\ is the mean number of surfaced decisions per day episode. Tools/dec.\ and Tokens/dec.\ summarize lookup-tool use and prompt-plus-completion tokens per surfaced decision.}

\begin{tabular}{lrrrrrr}
\toprule
\textbf{Model} & \textbf{Day clear} & \textbf{Final HP} & \textbf{Survivors} & \textbf{Decisions/ep.} & \textbf{Tools/dec.} & \textbf{Tokens/dec.} \\
\midrule
Gemini 3.1 Pro & $0.40$ & 0.58 & 0.80 & 328 & 1.27 & 124k \\
GPT-5.5 & $0.40$ & 0.38 & 0.55 & 391 & 0.77 & 114k \\
DeepSeek V4 & $0.20$ & 0.50 & 0.65 & 354 & 2.16 & 102k \\
Claude Sonnet 4.6 & $0.20$ & 0.50 & 0.60 & 354 & 0.91 & 70k \\
Grok 4.3 & $0.00$ & 0.20 & 0.20 & 361 & 0.63 & 163k \\
\bottomrule
\end{tabular}
\label{tab:day-summary}
\end{table}


The terminal traces show that losses are qualitatively different. Of the 76 failed Encounter episodes, 38 are party wipes, 25 end with the party essentially intact but failing an objective, or losing the race to a non-HP win condition, and 13 are depleted losses with intermediate remaining HP. \texttt{unseen\_predator} is the clearest wipe-driven failure: the party repeatedly fights a hidden, resistant enemy under unfavorable attack conditions. \texttt{west\_tower} is different: many losses occur before the party is meaningfully damaged because the opposing side secures the map objective first. \texttt{stonecut\_bridge} exposes a third pattern, where the party often remains alive but spends too many turns on low-value movement and attacks. These distinctions are important because a single binary loss does not say whether the model misunderstood the rules, ignored the objective, chose inefficient routes, or simply lost an attrition fight.

Table~\ref{tab:decision-space} gives statistics of the decision space that help to explain why these failures persist under full observability. Objective encounters have the longest average horizon, while Boss and Objective encounters expose the largest local option sets. The median decision is much smaller than the tail, but some decisions expose hundreds of legal choices and the maximum option count exceeds 1{,}000. The model therefore has to solve two problems at once: choose a good local commitment from a large legal set, and understand how that commitment advances the scenario-level objective.

The Day track compounds these pressures. \texttt{kestrel\_approach} is the cleanest example: every model fails the day while preserving the party, indicating an objective-recognition failure rather than a resource-depletion failure. Other days show the complementary pattern, where models win several encounters before the chain fails near the end. This is the intended distinction between the tracks. Encounter performance measures local tactical competence; Day performance measures whether that competence remains disciplined when hit points, spell slots, rests, and objective progress carry forward.

\begin{table}[t]
\centering
\small
\caption{Decision-space characterization of the Encounter track. Decisions/ep.\ is the mean number of surfaced decisions per episode. Options/dec.\ is the mean number of legal candidates exposed at each decision point, and Max options is the largest legal option set observed in the family. Objective encounters have the longest average horizons, while both Objective and Boss encounters expose large local choice sets.}
\begin{tabular}{lrrr}
\toprule
\textbf{Family} & \textbf{Decisions/ep.} & \textbf{Options/dec.} & \textbf{Max options} \\
\midrule
Duel & 52 & 67.7 & 169 \\
Objective & 112 & 47.3 & 1159 \\
Boss & 60 & 52.2 & 441 \\
\bottomrule
\end{tabular}
\label{tab:decision-space}
\end{table}

\section{Limitations}
DungeonBench scopes evaluation to tactical combat. It is designed for controlled comparison of rule-aware tactical decisions, not full tabletop play: social interaction, open-ended exploration, campaign-scale simulation, narrative adjudication, or broad party and build optimization. The main benchmark uses complete tactical observations and known encounter-day sequences. Strong DungeonBench performance should therefore be read as evidence about rule-aware tactical reasoning in this setting, not as a claim about all tabletop play or all long-horizon reasoning.

\textbf{Future extensions.}
Several natural extensions would test capabilities outside the current combat-focused benchmark. A \emph{Dungeon} track could chain multiple encounter days, extending the current Day setting to longer horizons. This would stress planning across long rests, spell preparation, consumable use, persistent injuries or afflictions, and the choice of which resources to commit when later days remain. Such a track could also introduce bounded noncombat choices, such as scouting, detours, stealth, and information-gathering spells like \emph{Arcane Eye}, \emph{Speak with Dead}, or \emph{Speak with Animals}, without requiring full open-ended campaign simulation. Hidden-information variants could restrict what agents know about enemy traits, unseen positions, future rooms, objectives, immunities, and other tactical facts until they are discovered or inferred. Multi-agent variants could assign separate policies to party members, monsters, or opposing sides, enabling coordination, adversarial play, self-play, and population-based training.

\section{Conclusion}
DungeonBench provides a rules-rich tactical-combat benchmark for evaluating agents under heterogeneous state-dependent decisions, long horizons, and compositional rules. Its core setting focuses on fully observed tactical decision-making, curated scenario suites, and engine-generated legal options. The benchmark asks whether agents can reason over tactical choices whose meaning depends on geometry, action economy, spell and item semantics, objectives, reactions, and persistent resources. The resulting task is deliberately close to the texture of tactical play: survive the current fight, advance the objective, respect the rules, and keep enough resources to matter later. DungeonBench makes those tactical commitments executable, repeatable, and measurable. Broader framework integrations beyond the current Gymnasium-compatible interface would make the same decision stream easier to use in external RL libraries and multi-agent experiments.

\bibliographystyle{plainnat}
\bibliography{references}

\clearpage

\appendix
\appendix
\begin{center}
	{\Large\bfseries Appendix}
\end{center}

\section{Policy Interface Details}
\label{app:policy-interface}

\subsection{Shared decision representation}
All policy implementations consume the same engine-generated decision stream. Each decision record contains a decision kind, acting combatant, structured tactical state, a finite list of legal options, and the selected option index. The state includes combatants, resources, positions, terrain, active effects, zones, summons, scene objects, initiative context, and encounter-day state. Options carry simulator-facing payloads together with policy-facing labels and normalized details such as targets, resource costs, movement facts, area parameters, or reaction triggers. This representation is the source for both text prompts and tensor features.

\subsection{Language-model policy}
The language-model controller receives a JSON prompt rather than free-form command instructions. The system instruction tells the model that the simulator has already generated the complete legal option list, that simulator state and legal options override outside assumptions, and that the model must return only JSON matching the response schema. The user prompt contains the current decision kind, actor, tactical state summary, optional encounter history, legal options, and a response schema such as \texttt{\{"choice\_id": "string option\_id"\}}.

The controller is invoked at each exposed decision point, not only at full-turn action selection: movement, turn control, reactions, optional riders, resource spends, and other follow-up decisions use the same choice-ID contract. For most decisions, legal options are rendered as indexed option records with labels and compact mechanics. Movement decisions use a compact destination list with optional route counts; when route consequences may matter, the model can request a route lookup that reveals path geometry, movement cost, opportunity attacks, hazards, zone triggers, readied reactions, and route-specific mechanics. Similar lookup tools expose relevant combatant mechanics when an ability, spell, feature, item, or condition could change the choice. Invalid responses are retried with the validation error and the same response schema; if no valid choice is returned after the retry budget, the decision is recorded as an LLM failure and no illegal action is executed. Decision logs record prompt hashes, selected option identifiers, attempts, tool calls, token usage, and optionally the serialized prompts for audit.

\subsection{Behavior-cloned option ranker}
Behavior cloning trains on decision traces produced by reference policies. Training traces are generated from procedurally varied episodes rather than only from the fixed headline suite. The generator samples encounter and day families, party and enemy rosters, tier, item buckets, authored or generated maps, template variants, and opponent policies; generated maps are validity-filtered for safe spawns, path connectivity, reachable objectives, and bounded terrain hazards.

Each supervised example contains the shared policy observation, the legal option set, and the selected option index. The tensorized representation includes global state features and categories, state tokens, a spatial state board, per-option features and categories, per-option spatial boards, and an action mask for variable option counts. The BC-Attention model embeds the global state, state tokens, and options, applies a small Transformer encoder over the combined sequence, and scores each legal option with masked logits. Training uses cross-entropy on the selected option, with validation grouped by scenario or family to avoid measuring only repeated decisions from the same episode source.

The current tensor encoder represents each decision with fixed global features, variable state tokens, spatial boards, and variable legal-option tensors. The global state vector contains 192 numeric and hashed categorical features; state tokens are 224-dimensional records for combatants, effects, zones, objectives, scene objects, terrain, and encounter context; and the actor-centered state board is a 27-channel $41 \times 41$ grid. Each legal option has a 224-dimensional feature vector, eight categorical fields covering action and payload type, timing, target side, damage type, spell kind, and feature kind, and a 6-channel option board marking movement paths, destinations, targets, areas, walls, and objects. Variable option sets are stored in packed form with offsets for trace datasets, padded only in batches or Gymnasium observations, and accompanied by an action mask so the policy scores only legal choices.

\subsection{PPO interface}
The PPO interface uses the same event-level decision source through a Gymnasium-compatible wrapper. The environment exposes the tensorized policy observation, a discrete action over padded option slots, and an action mask; invalid option indices are rejected before simulator execution. The actor reuses the BC variable-option policy architecture, and the critic adds a value head over the structured state and spatial board. Fine-tuning can initialize the actor from the behavior-cloned checkpoint and optimize sparse benchmark rewards with generalized advantage estimation and clipped PPO updates. Because PPO acts at the same decision points as the BC and LLM agents, fine-tuning changes option valuation rather than the benchmark interface.

\subsection{Reward and termination details}
Encounter episodes run from the initial state until victory, defeat, draw, or round-limit termination, with sparse learning reward $+1$ for victory, $-1$ for loss, $0$ for draw, and $-0.001$ per decision step. Day episodes run across linked encounters until full clear, party defeat, or scenario end, with reward $+1$ for clearing the full day, $-1$ for terminal party wipe, $+0.1$ per encounter cleared, and the same small step penalty. Raw returns are interpreted within a track rather than compared directly across Encounter and Day; damage, healing, resource use, survival margin, objective progress, and round efficiency are reported as metrics or trace diagnostics.

\section{Language-Model Prompt Format}
\label{app:llm-prompt-format}

Language-model policies interact with DungeonBench through a structured choice interface. The simulator has already enumerated the legal choices for the current decision; the model is asked to rank those choices and return a JSON object that identifies one legal option. This appendix describes the prompt contract used by the reported language-model policies. Long state summaries and legal option lists are shortened in examples, but the field names and response schemas match the evaluation interface.

\subsection{Message Structure}

Each language-model decision contains a fixed system instruction, a JSON user prompt, and a set of optional lookup tools. The final model message must be a JSON object matching the \texttt{response\_schema} in the user prompt. Tool calls may be made before the final message, but they do not replace the final choice response.

\begin{quote}
\small
You are controlling a combatant in a D\&D 2014-style tactical combat benchmark simulator. The engine has already generated the complete legal option list for this decision. Use your understanding of D\&D tactics, action economy, positioning, conditions, resources, and monster roles, but obey the simulator state and legal option list over outside assumptions. If a creature's ability, spell, feature, item, or condition mechanic could change your decision, prefer to look it up using the available tools rather than guessing. Choose exactly one \texttt{option\_id} from \texttt{legal\_options}. Do not invent actions, targets, movement, or rules. Return only a JSON object matching \texttt{response\_schema}.
\end{quote}

\subsection{User Prompt Fields}

The user prompt is a JSON object. Table~\ref{tab:llm-prompt-fields} lists all top-level fields used by the prompt contract. Only \texttt{decision}, \texttt{state}, \texttt{legal\_options}, and \texttt{response\_schema} are present on every first attempt; the remaining fields are conditional.

\begin{table}[h]
\centering
\small
\setlength{\tabcolsep}{4pt}
\caption{Top-level fields in the language-model user prompt. Conditional fields are included only when the corresponding decision family or retry path needs them.}
\begin{tabular}{p{0.28\linewidth}p{0.64\linewidth}}
\toprule
\textbf{Field} & \textbf{Description} \\
\midrule
\texttt{response\_schema} & Required JSON response shape for the current decision. The common case is \texttt{\{"choice\_id":"string option\_id"\}}. Specialized decisions add fields such as \texttt{path\_id}, target-id arrays, healing allocations, or wall geometry. \\
\texttt{decision} & Decision metadata: \texttt{kind} and the acting combatant reference. The actor reference contains \texttt{id}, \texttt{name}, \texttt{side}, and \texttt{creature\_type}; it is \texttt{null} for side-level or intermission decisions with no single actor. \\
\texttt{state} & Tactical state visible to the acting side, including round, active turn, combatants, resources, positions, objectives, battlefield, effects, zones, summons, scene objects, and compact rule/item indexes. \\
\texttt{encounter\_history} & Public event-history strings available to the acting side. This field is present when history prompting is enabled, as in the reported evaluations. \\
\texttt{legal\_options} & The finite legal option set produced by the simulator. The model must choose from this set or fill a specialized submit schema tied to a legal \texttt{submit} option. \\
\texttt{legal\_options\_schema} & Compact column schema for list-valued option encodings, used primarily for movement destinations. \\
\texttt{context} & Decision-specific auxiliary facts that are not naturally part of the global state, such as remaining movement, save DCs, selected coordinates, candidate target pools, or spell names. \\
\texttt{movement\_route\_lookup} & Present only on movement decisions. It explains that destination rows are compact and that \texttt{lookup\_movement\_routes} can reveal route geometry and route-specific consequences. \\
\texttt{previous\_errors} & Present only on retry attempts after a provider or validation failure. It records previous validation errors. \\
\texttt{repair\_instruction} & Present only on retry attempts. It asks the model to return only JSON matching \texttt{response\_schema}. \\
\bottomrule
\end{tabular}
\label{tab:llm-prompt-fields}
\end{table}

\subsection{State Summary Fields}

The \texttt{state} object is a compact observation rather than a full simulator dump. It contains enough public tactical information to rank legal options while exposing detailed rule text through lookup tools. The main state fields are:

\begin{itemize}
\item \texttt{round\_number} and \texttt{active\_turn\_actor\_id}: current turn context.
\item \texttt{turn\_spellcasting}: spellcasting restrictions and whether the acting creature has already cast a bonus-action or non-action cantrip spell this turn.
\item \texttt{scenario\_context}: encounter id, split, track, tags, difficulty, party level, objective summary, map metadata, and XP-difficulty metadata.
\item \texttt{objective\_progress}, \texttt{encounter\_flags}, and \texttt{side\_flags}: progress toward non-defeat objectives and scenario flags.
\item \texttt{combatants}: one record per visible or relevant combatant, including id, name, side, type, level/CR, armor class, hit points, position or last-known visibility payload, initiative, conditions, life/consciousness state, reactions, legendary resources, size, movement speeds, spell slots, feature uses, resource pools, equipped items, and visible pairwise distances.
\item \texttt{visible\_rule\_index}: compact per-combatant index of attacks, spells, features, traits, combat riders, senses, mobility, and defenses. This index supplies names for lookup tools.
\item \texttt{visible\_item\_index}: compact per-combatant index of visible carried or equipped items. This index supplies names for item lookup.
\item \texttt{battlefield}: ceiling height and relevant terrain cells, including blocking, difficult terrain, cover, hazards, elevation, gaps, liquids, climbability, and terrain tags.
\item \texttt{active\_effects}: ongoing effects with source, target, concentration, expiration, position, originating ability, conditions, semantic tags, and effect state.
\item \texttt{active\_zones}: active area or zone effects with source, position, template, blocking/cover/obscurement, damage, saves, trigger timing, magic suppression, projectile blocking, movement behavior, height, and state.
\item \texttt{active\_summons}: compact records for active summoned entities.
\item \texttt{scene\_objects}: interactive or attackable objects, their positions, hit points, available interactions, flags set by interactions, save/damage metadata, and object state.
\item \texttt{recent\_events}: compact recent engine events when separate \texttt{encounter\_history} prompting is disabled.
\end{itemize}

\subsection{Legal Option Encodings}

DungeonBench uses several legal-option encodings, all grounded in simulator-generated legality.

\paragraph{Structured option records.}
Most decisions render \texttt{legal\_options} as objects with an \texttt{option\_id}, a human-readable \texttt{label}, and decision-specific payload fields. Turn actions, reactions, spell defenses, feature modes, and target choices use this format. Action-like options may include \texttt{plan\_kind}, \texttt{payload\_kind}, \texttt{payload\_name}, \texttt{target}, \texttt{resource\_timing}, movement/template coordinates, execution mode, attack-sequence mode, and a compact \texttt{mechanics} object.

\paragraph{Compact movement rows.}
Movement decisions can contain hundreds of legal destinations, so they use compact rows:

\begin{quote}
\small
\begin{verbatim}
"legal_options_schema": [
  "option_id", "x", "y", "z", "cost_feet", "routes_if_multiple"
],
"legal_options": [
  ["d0", -2,  0, 0,  0],
  ["d1", -3, -1, 0,  5],
  ["d16", -2, 2, 0, 10, 2]
]
\end{verbatim}
\end{quote}

The optional final column indicates that a destination has multiple route realizations. If the selected destination has multiple routes, the final response must include a \texttt{path\_id} obtained from route lookup.

\paragraph{Boolean choices.}
Binary timing/resource decisions expose two legal options:

\begin{quote}
\small
\begin{verbatim}
[
  {"option_id": "decline", "label": "Decline", "choice": false},
  {"option_id": "use", "label": "Use Legendary Resistance ...", "choice": true}
]
\end{verbatim}
\end{quote}

\paragraph{Axis choices.}
Point and direction selection are decomposed into legal axis choices. For example, \texttt{point\_selection:x} first chooses an x-coordinate, then \texttt{point\_selection:y} chooses among y-values compatible with that x-coordinate, and \texttt{point\_selection:z} completes the point. Area direction uses the same pattern for \texttt{dx}, \texttt{dy}, and \texttt{dz}.

\paragraph{Submit choices.}
Some decisions require filling a structured allocation rather than selecting one action from many. These prompts expose a single legal option with \texttt{option\_id} equal to \texttt{submit}; the response must include \texttt{"choice\_id":"submit"} plus the allocation fields specified by \texttt{response\_schema}.

\subsection{Decision Families and Response Schemas}

The interface covers all policy decision kinds surfaced to the language model. The following families are exhaustive for the reported LLM controller.

\paragraph{Standard single-choice decisions.}
These decisions return one legal \texttt{option\_id}:

\begin{quote}
\small
\begin{verbatim}
{"choice_id": "option_0"}
\end{verbatim}
\end{quote}

This response shape is used for \texttt{turn\_control}, \texttt{turn\_action}, \texttt{free\_action:<timing>}, \texttt{turn\_opener}, \texttt{reaction}, \texttt{on\_hit\_rider}, \texttt{optional\_roll\_bonus:<kind>}, \texttt{before\_first\_attack}, \texttt{failed\_save\_success}, \texttt{targeted\_spell\_defense}, \texttt{dispel\_magic\_choice}, \texttt{redirected\_attack\_target}, \texttt{follow\_up\_attack\_target}, \texttt{feature\_mode}, and \texttt{intermission\_prep}. These prompts may add context such as target references, attack names, spell names, source names, roll totals, thresholds, short-rest state, or the available intermission combatants.

\paragraph{Boolean resource and timing decisions.}
These also return one legal \texttt{choice\_id}, but the legal option set is always \texttt{decline} versus \texttt{use}. This family includes \texttt{save\_reroll}, \texttt{legendary\_resistance}, \texttt{willing\_save\_failure}, \texttt{stroke\_of\_luck}, \texttt{relentless\_rage}, \texttt{relentless\_endurance}, \texttt{avert\_gaze}, \texttt{short\_rest}, and \texttt{hit\_die\_spend}. Context fields provide the relevant DC, total, natural roll, damage, damage type, source, recommendation flag, remaining hit dice, or remaining rests.

\begin{quote}
\small
\begin{verbatim}
{"choice_id": "use"}
{"choice_id": "decline"}
\end{verbatim}
\end{quote}

\subsection{Representative Legal Options}
\label{app:representative-legal-options}

The central abstraction in DungeonBench is the legal option list. At each surfaced decision point, the simulator
enumerates all currently legal choices and assigns stable identifiers; the policy selects among those identifiers rather
than issuing free-form commands. This makes the interface auditable and separates tactical valuation from command
parsing.

A movement decision is encoded compactly because the number of reachable destinations can be large. The following
excerpt is from a simple fighter-versus-goblin encounter; the full movement list contains many more destinations.

\begin{quote}
\small
\begin{verbatim}
"legal_options_schema": [
  "option_id", "x", "y", "z", "cost_feet", "routes_if_multiple"
],
"legal_options": [
  ["d0", -2,  0, 0,  0],
  ["d1", -3, -1, 0,  5],
  ["d2", -3,  0, 0,  5],
  ["d3", -3,  1, 0,  5],
  ["d4", -2, -1, 0,  5],
  ["d5", -2,  1, 0,  5],
  ["d6", -1, -1, 0,  5],
  ["d7", -1,  0, 0,  5]
]
\end{verbatim}
\end{quote}

Turn-action decisions use structured option records. The following abridged list shows the diversity of choices exposed
in a single decision: weapon attacks, generic combat actions, spell casts, an objective interaction, and explicit turn-
control options.

\begin{quote}
\small
\begin{verbatim}
"legal_options": [
  {
    "option_id": "option_0",
    "label": "action attack Mace -> Androsphinx",
    "slot_timing": "action",
    "plan_kind": "attack",
    "payload_kind": "attack",
    "payload_name": "Mace",
    "target": {"id": "right:androsphinx:1", "name": "Androsphinx"},
    "mechanics": {
      "attack_bonus": 6,
      "damage_formula": "1d6+2",
      "damage_type": "bludgeoning"
    }
  },
  {
    "option_id": "option_2",
    "label": "action dash -> Androsphinx",
    "slot_timing": "action",
    "plan_kind": "dash",
    "payload_kind": "none"
  },
  {
    "option_id": "option_10",
    "label": "action spell Sacred Flame -> Androsphinx",
    "slot_timing": "action",
    "plan_kind": "spell",
    "payload_kind": "spell",
    "payload_name": "Sacred Flame",
    "mechanics": {
      "kind": "save",
      "save_ability": "dex",
      "save_dc": 15,
      "damage_formula": "2d8",
      "damage_type": "radiant"
    }
  },
  {
    "option_id": "option_16",
    "label": "bonus spell Sanctuary -> Sigil Keeper",
    "slot_timing": "bonus",
    "plan_kind": "spell",
    "payload_kind": "spell",
    "payload_name": "Sanctuary",
    "mechanics": {
      "kind": "effect",
      "level": 1,
      "requires_concentration": false
    }
  },
  {
    "option_id": "option_15",
    "label": "action use object -> Sigil Keeper",
    "slot_timing": "action",
    "plan_kind": "use_object",
    "payload_kind": "objectinteractionplan"
  },
  {
    "option_id": "option_20",
    "label": "skip Action",
    "control_kind": "skip_action",
    "slot_timing": "action"
  },
  {
    "option_id": "option_22",
    "label": "end turn",
    "control_kind": "end_turn"
  }
]
\end{verbatim}
\end{quote}

Feature options use the same structured record format. For example, a bonus-action feature may appear as:

\begin{quote}
\small
\begin{verbatim}
{
  "option_id": "option_0",
  "label": "bonus feature Step of the Wind -> Ren Sable",
  "slot_timing": "bonus",
  "plan_kind": "feature",
  "payload_kind": "feature",
  "payload_name": "Step of the Wind",
  "mechanics": {
    "kind": "effect",
    "requires_concentration": false
  }
}
\end{verbatim}
\end{quote}

The policy response is intentionally small relative to the option list. For the turn-action example above, selecting the
objective interaction is simply:

\begin{quote}
\small
\begin{verbatim}
{"choice_id": "option_15"}
\end{verbatim}
\end{quote}

\paragraph{Movement decisions.}
Movement responses choose a compact destination option. If the destination has exactly one route, \texttt{path\_id} may be omitted. If the destination has multiple legal route realizations, \texttt{path\_id} must identify the route selected from \texttt{lookup\_movement\_routes}.

\begin{quote}
\small
\begin{verbatim}
{"choice_id": "d77"}
{"choice_id": "d16", "path_id": "path_1"}
\end{verbatim}
\end{quote}

\paragraph{Coordinate and direction decisions.}
Point and direction decisions are exposed as sequential axis decisions. Each step still returns a single legal \texttt{choice\_id}. A point-selection x-step may include a legal done option once the minimum number of points has been selected.

\begin{quote}
\small
\begin{verbatim}
{"choice_id": "point_selection:x|x 4"}
{"choice_id": "point_selection:y|y -1"}
{"choice_id": "point_selection:z|z 0"}
{"choice_id": "point_selection:x|done selecting points"}

{"choice_id": "area_direction:dx|dx 1"}
{"choice_id": "area_direction:dy|dy 0"}
{"choice_id": "area_direction:dz|dz 0"}
\end{verbatim}
\end{quote}

\paragraph{Area protection.}
Area-protection prompts are used for abilities such as careful or sculpted spell placement. The prompt context includes the spell, targets in resolution order, careful candidates, sculpt candidates, and capacities. The response submits both target-id lists.

\begin{quote}
\small
\begin{verbatim}
{
  "choice_id": "submit",
  "careful_target_ids": ["left:cleric:1"],
  "sculpt_target_ids": ["left:fighter:1", "left:rogue:1"]
}
\end{verbatim}
\end{quote}

\paragraph{Multi-target allocation.}
Multi-target allocation prompts distribute a fixed number of effect packets across a candidate pool. Context contains \texttt{spell}, \texttt{primary\_target}, \texttt{recipient\_pool}, \texttt{packet\_count}, and \texttt{allow\_repeat}. The response has exactly \texttt{packet\_count} target ids, with repetition allowed only when \texttt{allow\_repeat} is true.

\begin{quote}
\small
\begin{verbatim}
{
  "choice_id": "submit",
  "packet_target_ids": [
    "right:orc:1",
    "right:orc:2",
    "right:orc:2"
  ]
}
\end{verbatim}
\end{quote}

\paragraph{Healing pool allocation.}
Healing pool prompts allocate a bounded healing pool among candidate recipients. Context contains \texttt{spell}, \texttt{primary\_target}, \texttt{recipient\_pool}, and \texttt{total\_pool}. The response is an array of unique target allocations with nonnegative integer amounts whose total does not exceed the pool.

\begin{quote}
\small
\begin{verbatim}
{
  "choice_id": "submit",
  "healing_allocations": [
    {"target_id": "left:fighter:1", "amount": 8},
    {"target_id": "left:wizard:1", "amount": 4}
  ]
}
\end{verbatim}
\end{quote}

\paragraph{Unaffected targets.}
Unaffected-target prompts choose creatures to exclude from an effect. Context contains the spell and a candidate pool. The response is a unique array drawn from that pool.

\begin{quote}
\small
\begin{verbatim}
{
  "choice_id": "submit",
  "unaffected_target_ids": ["left:paladin:1"]
}
\end{verbatim}
\end{quote}

\paragraph{Wall path.}
Wall-path prompts choose a compact wall geometry. Context contains \texttt{spell}, \texttt{anchor\_point}, \texttt{target\_position}, \texttt{panel\_count}, and \texttt{allowed\_panel\_styles}. The response specifies the style, initial cardinal direction, three segment lengths summing to \texttt{panel\_count}, and two turns.

\begin{quote}
\small
\begin{verbatim}
{
  "choice_id": "submit",
  "panel_style": "line",
  "initial_direction": "north",
  "segment_lengths": [2, 2, 1],
  "turns": ["left", "straight"]
}
\end{verbatim}
\end{quote}

\subsection{Decision-Specific Context Fields}

The \texttt{context} object is omitted when no auxiliary fields are needed. When present, its fields are as follows:

\begin{itemize}
\item \texttt{movement}: \texttt{movement\_remaining\_feet}.
\item \texttt{on\_hit\_rider}: \texttt{target}, \texttt{rider\_name}.
\item \texttt{optional\_roll\_bonus:<kind>}: \texttt{roll\_total}, \texttt{threshold}.
\item \texttt{before\_first\_attack}: \texttt{target}, \texttt{attack\_name}, \texttt{attack\_mode}.
\item \texttt{save\_reroll}: \texttt{trait\_name}, \texttt{ability}, \texttt{total}, \texttt{dc}.
\item \texttt{legendary\_resistance}: \texttt{ability}, \texttt{dc}, \texttt{source\_name}.
\item \texttt{willing\_save\_failure}: \texttt{source}, \texttt{spell}, \texttt{reason}, \texttt{recommended}.
\item \texttt{failed\_save\_success}: \texttt{source\_name}.
\item \texttt{targeted\_spell\_defense}: \texttt{spell}.
\item \texttt{dispel\_magic\_choice}: \texttt{spell}, \texttt{target}, \texttt{target\_name}.
\item \texttt{area\_protection}: \texttt{spell}, \texttt{targets\_in\_order}, \texttt{careful\_candidates}, \texttt{sculpt\_candidates}.
\item \texttt{multi\_target\_allocation}: \texttt{spell}, \texttt{primary\_target}, \texttt{recipient\_pool}, \texttt{packet\_count}, \texttt{allow\_repeat}.
\item \texttt{healing\_pool\_allocation}: \texttt{spell}, \texttt{primary\_target}, \texttt{recipient\_pool}, \texttt{total\_pool}.
\item \texttt{point\_selection:x/y/z}: \texttt{spell}, \texttt{axis}, \texttt{selected\_x}, \texttt{selected\_y}, \texttt{selected\_points}, \texttt{minimum\_count}, \texttt{maximum\_count}, \texttt{allow\_repeat}, \texttt{require\_connected}, \texttt{adjacency\_size\_feet}.
\item \texttt{area\_direction:dx/dy/dz}: \texttt{spell}, \texttt{axis}, \texttt{selected\_dx}, \texttt{selected\_dy}, \texttt{candidate\_direction\_count}.
\item \texttt{unaffected\_targets}: \texttt{spell}, \texttt{candidate\_pool}.
\item \texttt{wall\_path}: \texttt{spell}, \texttt{anchor\_point}, \texttt{target\_position}, \texttt{panel\_count}, \texttt{allowed\_panel\_styles}.
\item \texttt{redirected\_attack\_target}: \texttt{hostile\_actor}, \texttt{feature\_name}, \texttt{attack\_name}.
\item \texttt{follow\_up\_attack\_target}: \texttt{source\_name}, \texttt{attack\_name}.
\item \texttt{feature\_mode}: \texttt{target}.
\item \texttt{stroke\_of\_luck}: \texttt{total}, \texttt{threshold}, \texttt{natural\_roll}.
\item \texttt{relentless\_rage}: \texttt{trait\_name}, \texttt{dc}.
\item \texttt{relentless\_endurance}: \texttt{trait\_name}, \texttt{damage}, \texttt{damage\_type}.
\item \texttt{avert\_gaze}: \texttt{source}, \texttt{source\_name}, \texttt{recommended}.
\item \texttt{short\_rest}: \texttt{rests\_remaining}, \texttt{short\_rest\_recovery\_available}, \texttt{combatants}.
\item \texttt{hit\_die\_spend}: \texttt{remaining\_hit\_dice}, \texttt{hit\_die\_sides}, \texttt{missing\_hit\_points}.
\item \texttt{intermission\_prep}: \texttt{short\_rest\_taken}, \texttt{short\_rest\_available}, \texttt{combatants}.
\end{itemize}

\subsection{Lookup Tools}

The model receives lookup tools listed in \ref{tab:llm-lookup-tools} alongside the prompt. These tools expose detailed mechanics while keeping the prompt compact and preserving a clear separation between visible summaries and full rule payloads. The final response remains a choice JSON after any tool calls.

\begin{table}[h]
\centering
\small
\setlength{\tabcolsep}{4pt}
\caption{Lookup tools available to language-model policies. Names for combatant capabilities come from \texttt{state.visible\_rule\_index}; item names come from \texttt{state.visible\_item\_index}; movement route ids come from \texttt{legal\_options}.}
\begin{tabular}{p{0.28\linewidth}p{0.60\linewidth}}
\toprule
\textbf{Tool} & \textbf{Arguments and returned information} \\
\midrule
\texttt{lookup\_spell} & Arguments: \texttt{combatant\_id}, \texttt{spell\_name}. Returns full spell mechanics, including slot cost, range, target constraints, attack/save mode, damage or healing, area/zone/summon details, concentration, and effects. \\
\texttt{lookup\_feature} & Arguments: \texttt{combatant\_id}, \texttt{feature\_name}. Returns class, monster, item-granted, or effect-granted feature mechanics. \\
\texttt{lookup\_attack} & Arguments: \texttt{combatant\_id}, \texttt{attack\_name}. Returns attack bonus, damage, damage type, reach/range, target constraints, magical or silvered properties, and on-hit effects. \\
\texttt{lookup\_combat\_rider} & Arguments: \texttt{combatant\_id}, \texttt{rider\_name}. Returns on-hit rider triggers, eligibility, extra damage, saves, conditions, forced movement, healing, or follow-up consequences. \\
\texttt{lookup\_trait} & Arguments: \texttt{combatant\_id}, \texttt{trait\_name}. Returns passive traits, senses, mobility, immunities, advantage rules, special targeting or movement permissions, and other always-on mechanics. \\
\texttt{lookup\_item} & Arguments: \texttt{combatant\_id}, \texttt{item\_name}. Returns visible item state and item mechanics, including attacks, granted features or spells, passive effects, charges, attunement/equipped state, damage thresholds, curses, and stored spells. \\
\texttt{lookup\_condition} & Argument: \texttt{name}. Returns the benchmark definition of a condition such as \texttt{Prone}, \texttt{Restrained}, or \texttt{Frightened}. \\
\texttt{lookup\_movement\_routes} & Movement decisions only. Argument: \texttt{option\_id}. Returns path geometry, costs, opportunity attacks, hazards, zone triggers, readied reactions, and other route-specific mechanics for a legal movement destination. \\
\bottomrule
\end{tabular}
\label{tab:llm-lookup-tools}
\end{table}

The tool declarations follow the provider's function/tool schema. The exact provider wrapper differs across APIs, but the logical declaration has the following form:

\begin{quote}
\small
\begin{verbatim}
[
  {
    "name": "lookup_spell",
    "description": "Get full mechanics of a spell a combatant can cast...",
    "parameters": {
      "type": "object",
      "properties": {
        "combatant_id": {"type": "string"},
        "spell_name": {"type": "string"}
      },
      "required": ["combatant_id", "spell_name"]
    }
  },
  {
    "name": "lookup_movement_routes",
    "description": "Get route geometry and consequences for a movement destination...",
    "parameters": {
      "type": "object",
      "properties": {
        "option_id": {"type": "string"}
      },
      "required": ["option_id"]
    }
  }
]
\end{verbatim}
\end{quote}

Combatant lookup tools return JSON with the following shape:

\begin{quote}
\small
\begin{verbatim}
{
  "combatant_id": "left:wizard:1",
  "combatant_name": "Wizard",
  "lookup_kind": "spell",
  "lookup_index": 3,
  "name": "Fireball",
  "payload": {
    "...": "full mechanics elided"
  }
}
\end{verbatim}
\end{quote}

Movement route lookup returns destination-level route details:

\begin{quote}
\small
\begin{verbatim}
{
  "option_id": "d16",
  "label": "move to (-2,2,0)",
  "routes": 2,
  "payload": {
    "to": [-2, 2, 0],
    "cost": 10,
    "path_options": [
      {
        "path_id": "path_0",
        "cost": 10,
        "path": [[-2,0,0], [-2,1,0], [-2,2,0]],
        "mechanics": {"...": "route consequences elided"}
      }
    ]
  }
}
\end{verbatim}
\end{quote}

  \subsection{Tool-Call Example: Route-Sensitive Movement}
  \label{app:route-lookup-example}

  Movement destinations are listed compactly in the prompt, but a destination can have multiple legal routes with
  different tactical consequences. In that case, the model may inspect route details before selecting a final movement
  response.

  \begin{quote}
  \small
  \begin{verbatim}
  "legal_options": [
    ["d0", -2,  0, 0,  0],
    ["d16", -2, 2, 0, 10, 2]
  ]
  \end{verbatim}
  \end{quote}

  The final column indicates that \texttt{d16} has two legal route realizations. A model can call the route lookup tool:

  \begin{quote}
  \small
  \begin{verbatim}
  lookup_movement_routes({"option_id": "d16"})
  \end{verbatim}
  \end{quote}

  The tool returns route-specific geometry and consequences:

  \begin{quote}
  \small
  \begin{verbatim}
  {
    "option_id": "d16",
    "label": "move to (-2,2,0)",
    "routes": 2,
    "payload": {
      "to": [-2, 2, 0],
      "cost": 10,
      "path_options": [
        {
          "path_id": "path_0",
          "cost": 10,
          "path": [[-2,0,0], [-2,1,0], [-2,2,0]],
          "mechanics": {
            "opportunity_attacks": [],
            "hazards": []
          }
        },
        {
          "path_id": "path_1",
          "cost": 10,
          "path": [[-2,0,0], [-1,1,0], [-2,2,0]],
          "mechanics": {
            "opportunity_attacks": ["right:goblin:1"],
            "hazards": []
          }
        }
      ]
    }
  }
  \end{verbatim}
  \end{quote}

  The final answer then selects both the destination and the route:

  \begin{quote}
  \small
  \begin{verbatim}
  {"choice_id": "d16", "path_id": "path_0"}
  \end{verbatim}
  \end{quote}

  This example illustrates why movement lookup is separate from the initial prompt: most destination choices can be
  represented compactly, while route-sensitive cases can still expose opportunity attacks, hazards, zone triggers, and
  other route-specific consequences when they matter.

\subsection{Movement Prompt}

The following excerpt shows a movement decision from a simple duel. The full decision contains 168 legal destinations; only the first few are shown.

\begin{quote}
\small
\begin{verbatim}
{
  "response_schema": {
    "choice_id": "string destination option_id",
    "path_id": "string path_id from the selected option's path_options; 
    omit when the selected option does not include path_options"
  },
  "decision": {
    "kind": "movement",
    "actor": {
      "id": "left:fighter:1",
      "name": "Fighter",
      "side": "left",
      "creature_type": "humanoid"
    }
  },
  "state": {
    "round_number": 1,
    "active_turn_actor_id": "left:fighter:1",
    "combatants": "...",
    "visible_rule_index": "...",
    "visible_item_index": "..."
  },
  "encounter_history": [
    "Round 1 begins."
  ],
  "legal_options_schema": [
    "option_id", "x", "y", "z", "cost_feet", "routes_if_multiple"
  ],
  "legal_options": [
    ["d0", -2,  0, 0,  0],
    ["d1", -3, -1, 0,  5],
    ["d2", -3,  0, 0,  5],
    ["d3", -3,  1, 0,  5],
    ["d4", -2, -1, 0,  5],
    ["d5", -2,  1, 0,  5]
  ],
  "context": {
    "movement_remaining_feet": 30
  },
  "movement_route_lookup_instruction": "..."
}
\end{verbatim}
\end{quote}

\subsection{Action Prompt}

Turn-action prompts use structured option records. The excerpt below shows the first options from a fighter's action decision against a goblin.

\begin{quote}
\small
\begin{verbatim}
{
  "response_schema": {
    "choice_id": "string option_id"
  },
  "decision": {
    "kind": "turn_action",
    "actor": {
      "id": "left:fighter:1",
      "name": "Fighter",
      "side": "left",
      "creature_type": "humanoid"
    }
  },
  "state": "...",
  "encounter_history": [
    "Round 1 begins.",
    "Round 1: Fighter moves 20 feet."
  ],
  "legal_options": [
    {
      "option_id": "option_0",
      "label": "action attack Longsword -> Goblin",
      "slot_timing": "action",
      "plan_kind": "attack",
      "payload_kind": "attack",
      "payload_name": "Longsword",
      "target": {
        "id": "right:goblin:1",
        "name": "Goblin",
        "side": "right",
        "creature_type": "humanoid"
      },
      "resource_timing": "action",
      "mechanics": {
        "attack_bonus": 5,
        "damage_formula": "1d8+3",
        "damage_type": "slashing"
      }
    },
    {
      "option_id": "option_1",
      "label": "action attack Unarmed Strike -> Goblin",
      "payload_kind": "attack",
      "payload_name": "Unarmed Strike"
    },
    {
      "option_id": "option_2",
      "label": "action dash -> Goblin",
      "plan_kind": "dash"
    }
  ]
}
\end{verbatim}
\end{quote}

\subsection{Retry Prompt}

If a model response is not parseable JSON, omits \texttt{choice\_id}, selects an illegal id, omits a required specialized field, or violates a submit-allocation validator, the next prompt repeats the same decision and adds repair fields:

\begin{quote}
\small
\begin{verbatim}
{
  "response_schema": {"choice_id": "string option_id"},
  "decision": "...",
  "state": "...",
  "encounter_history": "...",
  "legal_options": "...",
  "previous_errors": [
    "choice_id 'option_99' is not one of ['option_0', 'option_1']"
  ],
  "repair_instruction": "One or more previous responses failed validation or
  provider delivery. Return only JSON matching response_schema."
}
\end{verbatim}
\end{quote}

\clearpage
\section{Encounter Scenario Catalog}
\begin{table}[h]
\centering
\scriptsize
\setlength{\tabcolsep}{3pt}
\renewcommand{\arraystretch}{1.06}
\caption{Encounter-track scenario catalog. These are the 20 standalone Encounter scenarios used in the main evaluation runs. Scenario IDs are stable evaluation identifiers; the Day-track table separately lists linked encounters inside each encounter-day scenario.}
\begin{tabular}{@{}>{\raggedright\arraybackslash}p{0.25\linewidth}>{\raggedright\arraybackslash}p{0.21\linewidth}>{\raggedright\arraybackslash}p{0.47\linewidth}@{}}
\toprule
\textbf{Scenario ID} & \textbf{Family / pressure} & \textbf{Tactical focus} \\
\midrule
\texttt{blackstar\_crucible} & Epic boss counterplay & Epic casters face a lich and guardian around sight breaks and hazards; counterspell, dispel, burst timing, and survival trade off. \\
\texttt{briar\_gate} & Zone movement & Roots, forced movement, chokepoints, and hazards make pathing and control placement decisive. \\
\texttt{captains\_crossing} & Martial mirror & Symmetric captain duel testing multiattack sequencing, range bands, and parry reactions. \\
\texttt{cinder\_nursery} & Swarm pressure & Clustered low-tier threats and vent hazards test whether agents treat small enemies as a tactical mass. \\
\texttt{copper\_cairn} & Material hazard & Corrosive oozes and ruin hazards test damage-type awareness, resistances, and safe objective pursuit. \\
\texttt{duel\_fighter\_goblin} & Martial baseline & Low-complexity duel covering movement, attacks, misses, and basic resource use. \\
\texttt{emberhall\_duel} & Caster duel & Compact high-agency spell fight testing burst timing, reaction use, and concentration pressure. \\
\texttt{glass\_maw} & Body control & Narrow terrain and engulfing bodies test spacing, rescue priorities, and restraint handling. \\
\texttt{kraken\_depths} & Environmental objective & Underwater movement, air-bell objectives, and a huge threat make environment planning central. \\
\texttt{lantern\_court} & Possession pressure & Incorporeal undead, possession recovery, radiant pressure, and control loss shape target priority. \\
\texttt{mirror\_hall} & Gaze objective & Pillars, cover, line of sight, and gaze pressure test objective commitment under monster-trait constraints. \\
\texttt{moonhook\_eyrie} & Vertical skirmish & Broken platforms, flying enemies, falling risk, and Feather Fall make vertical positioning tactical. \\
\texttt{nullglass\_chapel} & Affliction caster & Curses, cleansing, dispel choices, healing, and line-of-sight discipline drive survival. \\
\texttt{riftmill\_bazaar} & Summon objective & Summoners, mobile skirmishers, minion economy, and contested anchors test objective timing. \\
\texttt{rootcoil\_cistern} & Form rescue & Flooded terrain, grapples, restraint, swallowing, and form change test escape and rescue valuation. \\
\texttt{severance\_gallery} & Reaction control & Broken sightlines and opposing casters make reaction timing, movement, and spell selection central. \\
\texttt{sphinx\_sanctum} & Legendary lair & Legendary pressure, lair tempo, and a time-anchor objective test boss survival under a win-condition race. \\
\texttt{stonecut\_bridge} & Item objective & A talisman-bearing thief must balance artifact pressure, bridge control, mobility, and survival. \\
\texttt{unseen\_predator} & Visibility filter & Invisible, resistant opposition tests disadvantage management, hidden-target reasoning, and attrition discipline. \\
\texttt{west\_tower} & Objective pressure & Compact breach map where reinforcements, extraction, and map mutation punish static play. \\
\bottomrule
\end{tabular}
\label{tab:encounter-catalog}
\end{table}

\clearpage
\section{Encounter Trace Excerpt}
\label{app:trace-excerpt}
\subsection{Encounter Trace Excerpt}
\label{app:encounter-trace-excerpt}

The excerpt below is an abridged segment from a \texttt{moonhook\_eyrie}
run. This encounter is a useful diagnostic example because it combines
vertical positioning, flying enemies, forced route commitments, readied
attacks, falling mitigation, spell targeting, condition markers, and reactive
defenses within the same combat. 

\begin{quote}
\scriptsize
\begin{verbatim}
Round 1 begins.
Round 1: Air Elemental [right] readies Slam for the first hostile that enters reach.
Round 1: Talan Ashbow (Hunter Ranger 11) moves.
Round 1: Talan Ashbow uses Volley, but catches nobody.
Round 1: Talan Ashbow moves.
Round 1: Talan Ashbow ends their turn.
Round 1: Fireball's origin is blocked and appears at (6,-1,4) instead.
Round 1: Vael Tindersight (Evocation Wizard 11) casts Fireball, but catches nobody.
Round 1: Ren Sable (Open Hand Monk 11) moves.
Round 1: Ren Sable dashes toward Manticore.
Round 1: Ren Sable uses Step of the Wind.
Round 1: Ren Sable dashes toward Manticore.
Round 1: Manticore attacks Vael with Tail Spike (4 + 5 = 9) and misses.
Round 1: Manticore attacks Vael with Tail Spike (1 + 5 = 6) and misses.
Round 1: Manticore attacks Vael with Tail Spike (8 + 5 = 13) and misses.
Round 1: Vael casts Feather Fall on Aurel Dawn, slowing the descent.
Round 1: Aurel Dawn (Life Cleric 11) drifts 60 feet under Feather Fall.
Round 1: Aurel Dawn moves.
Round 1: Aurel Dawn's Guiding Bolt hits Wyvern for 18 radiant damage (92/110 HP left).
Round 1: Wyvern gains Guiding Bolt.
Round 1: Wyvern moves.
Round 1: Wyvern readies Claws for the first hostile that enters reach.

...

Round 3 begins.
Round 3: Air Elemental [right] readies Slam for the first hostile that enters reach.
Round 3: Talan hits Wyvern with Shortsword for 9 piercing damage (30/110 HP left).
Round 3: Guiding Bolt on Wyvern is spent.
Round 3: Talan hits Wyvern with Shortsword critical for 13 piercing damage (17/110 HP left).
Round 3: Talan hits Wyvern with Shortsword (Offhand) for 5 piercing damage (12/110 HP left).
Round 3: Vael moves.
Round 3: Vael casts Burning Hands, catching Wyvern.
Round 3: Wyvern fails the save against Vael's Burning Hands, taking 18 fire damage (0/110 HP left).
Round 3: Wyvern is defeated.
Round 3: Ren uses Step of the Wind.
Round 3: Ren dashes toward Air Elemental [right].
Round 3: Manticore attacks Vael with Tail Spike (2 + 5 = 7) and misses.
Round 3: Manticore hits Vael with Tail Spike for 4 piercing damage (75/79 HP left).
Round 3: Manticore hits Vael with Tail Spike for 9 piercing damage (66/79 HP left).
Round 3: Aurel Dawn moves.
Round 3: Aurel Dawn's Guiding Bolt hits Air Elemental [right] for 22 radiant damage (68/90 HP left).
Round 3: Air Elemental [right] gains Guiding Bolt.
Round 4 begins.
Round 4: Talan hits Manticore with Longbow for 6 piercing damage (29/68 HP left).
Round 4: Vael casts Magic Missile at 3rd-level; 5 darts strike Manticore for 20 force damage.
Round 4: Ren attacks Air Elemental [right] with Dart (3 + 9 = 12) and misses.
Round 4: Guiding Bolt on Air Elemental [right] is spent.
Round 4: Manticore attacks Vael with Tail Spike.
Round 4: Vael casts Shield.
Round 4: Manticore attacks Vael with Tail Spike (9 + 5 = 14) and misses.
Round 4: Manticore hits Vael with Tail Spike critical for 11 piercing damage (55/79 HP left).
Round 4: Aurel Dawn's Guiding Bolt hits Air Elemental [right] for 25 radiant damage (43/90 HP left).
Round 4: Air Elemental [right] gains Guiding Bolt.

...

\end{verbatim}
\end{quote}

\clearpage
\section{Day-Track Scenario Breakdown}
\label{app:day-track-breakdown}
\begin{table}[h]
\centering
\small
\setlength{\tabcolsep}{4pt}
\renewcommand{\arraystretch}{1.1}
\caption{Day-track scenario breakdown. Each day is listed with party level, short-rest budget, encounter order, opposing roster, and the tactical pressure tested by each encounter.}
\begin{tabular}{@{}c>{\raggedright\arraybackslash}p{0.16\linewidth}>{\raggedright\arraybackslash}p{0.27\linewidth}>{\raggedright\arraybackslash}p{0.43\linewidth}@{}}
\toprule
\textbf{\#} & \textbf{Encounter} & \textbf{Enemies} & \textbf{What it tests} \\
\midrule
\multicolumn{4}{@{}l}{\textit{Graywatch March} -- level-5 party, 1 short rest, low-tier day baseline} \\
\midrule
1 & North Nave & manticore, veteran & Pre-rest skirmish: airborne ranged pressure with a martial frontline before any recovery window. \\
2 & Lower Yard & knight, scout & Post-rest crossing on a small map; rewards target priority and clean spacing. \\
3 & Ridge Room & mage & Caster duel on depleted resources; spacing and target priority decide the close. \\
\addlinespace
\multicolumn{4}{@{}l}{\textit{Hollowmere Keep} -- level-11 party, 2 short rests, long-form dungeon day} \\
\midrule
1 & Old Gate & 2$\times$ veteran, mage & Guarded entrance with enough ranged pressure to punish a careless approach. \\
2 & Pillar Hall & medusa, roper, gargoyle & Cover, gaze, and a restraining brute pull the party in different tactical directions. \\
3 & Lower Works & stone golem, wight & Hard attrition fight built around a durable construct and an undead lieutenant. \\
4 & Well Room & mage, 2$\times$ water elemental & Caster plus elementals pressure area control after the second recovery window. \\
5 & High Roof & young red dragon, half-red-dragon veteran, cult fanatic & Finale combining flight, breath pressure, and supporting fire. \\
\addlinespace
\multicolumn{4}{@{}l}{\textit{Kestrel Approach} -- level-11 party, 1 short rest, objective day} \\
\midrule
1 & West Tower & phase summoner, anchor knight, rift hound & Mixed martial-caster team fighting around reinforcements and a compact objective map. \\
2 & Lower Seals & sigil keeper, blink duelist, wand slinger & Two casters and a wand bearer contest objects, sight lines, and action economy. \\
3 & Catwalk & tempest lictor & Lair-backed solo boss; an environmental lever is available but not mandatory. \\
\addlinespace
\multicolumn{4}{@{}l}{\textit{Crownfall Keep} -- level-20 party, 1 short rest, high-tier setpiece day} \\
\midrule
1 & Ash Yard & pit fiend & Single fiend pressuring damage, reach, and action economy. \\
2 & Bell Vault & black bell of Ordrune & Construct boss; the party must coordinate burst and control before the climb. \\
3 & High Roof & ancient red dragon & Ancient-dragon finale on broken high ground. \\
\addlinespace
\multicolumn{4}{@{}l}{\textit{Starfall Observatory} -- level-11 party, 1 short rest, layered rule-interaction day} \\
\midrule
1 & Moonhook Eyrie & wyvern, 2$\times$ air elemental, manticore & Vertical opener: airborne enemies and fall risk punish flat-map habits. \\
2 & Rootcoil Cistern & roper, chuul, giant shark, giant toad & Flooded fight where grapples, restraint, swallowing, and form-change survivability all matter. \\
3 & Riftmill Bazaar & phase summoner, 2$\times$ rift hound, blink duelist & Summoner skirmish; new bodies and anchor objectives compete with immediate damage. \\
4 & Nullglass Chapel & mummy lord & Affliction finale: cleanse, dispel, healing, and remaining high slots decide the finish. \\
\bottomrule
\end{tabular}

\label{tab:day-encounters}
\end{table}

\clearpage

\end{document}